\documentclass[final]{siamart1116}
\graphicspath{ {Img/} }

\ifpdf
\DeclareGraphicsExtensions{.eps,.pdf,.png,.jpg}
\else
\DeclareGraphicsExtensions{.eps}
\fi

\newcommand{\vc}{{\boldsymbol c}}
\newcommand{\vd}{{\boldsymbol d}}
\newcommand{\vv}{{\boldsymbol v}}

\newcommand{\bR}{\mathbb{R}}

\newcommand{\C}{{\mathbb C}}
\newcommand{\R}{{\mathbb R}}

\newcommand{\F}{\mathcal{F}}
\newcommand{\Hil}{\ensuremath{\mathcal{H}}}
\newcommand{\tspan}{{\text{span}}}

\newcommand{\etc}{{\it etc.}}
\newcommand{\f}{\tilde{f}}

\newcommand{\argmin}{\mathop{\rm argmin}}

\newcommand\globalwidth{0.4}

\usepackage{float}
\usepackage{multirow}
\usepackage{boldline}
\usepackage[caption=false]{subfig}
\usepackage{setspace}
\usepackage[noadjust]{cite}
\usepackage{amsfonts}
\usepackage{epstopdf}
\usepackage{color}
\usepackage{enumerate}
\usepackage{mathtools}
\usepackage{array}

\usepackage{algorithm}
\usepackage{algorithmic}
\usepackage{wasysym}

\newcommand*{\vpointer}{\vcenter{\hbox{\scalebox{2}{\Huge\pointer}}}}

\usepackage{cleveref}
\numberwithin{theorem}{section}

\def \cR {\mathcal R}

\headers{PCM-TV-TFV: A Novel Framework for Image Reconstruction}
{Weihong Guo, Guohui Song, and Yue Zhang}
\title{{PCM-TV-TFV: A Novel Two Stage Framework for Image Reconstruction from Fourier Data}\thanks{This work is supported in part by grants NSF-DMS-1521582, NSF-DMS-204609, and NSF-DMS-1521661.}}

\author{Weihong Guo%
	\thanks{Department of Mathematics, Applied Mathematics and Statistics, Case Western Reserve University, Cleveland, OH (\email{wxg49@case.edu, yxz772@case.edu}).}%
	\and
	Guohui Song%
	\thanks{Department of Mathematics, Clarkson University, Potsdam, NY
		(\email{gsong@clarkson.edu}).}
	\and
	Yue Zhang%
	\footnotemark[2]
}

\begin{document}

\maketitle

\begin{abstract}
We propose in this paper a novel two-stage Projection Correction Modeling (PCM) framework for image reconstruction from (non-uniform) Fourier measurements. PCM consists of a projection stage (P-stage) motivated by the multi-scale Galerkin method and a correction stage (C-stage) with an edge guided regularity fusing together the advantages of total variation (TV) and total fractional variation (TFV). The P-stage allows for continuous modeling of the underlying image of interest. The given measurements are projected onto a space in which the image is well represented.  We then enhance the reconstruction result at the C-stage that minimizes an energy functional consisting of a fidelity in the transformed domain and a novel edge guided regularity. We further develop efficient proximal algorithms to solve the corresponding optimization problem. Various numerical results in both 1D signals and 2D images have also been presented to demonstrate the superior performance of the proposed two-stage method to other classical one-stage methods.
\end{abstract}

\begin{keywords}
edge guided reconstruction, Fourier measurements, total fractional order variation.
\end{keywords}

\begin{AMS}
	35R11, 65K10, 65F22, 90C25
\end{AMS}

\section{Introduction}
Image reconstruction from Fourier measurements has been a fundamental problem in various applications, such as magnetic resonance imaging (MRI)\cite{lustig_sparse_2007,lustig_compressed_2008,pruessmann1999sense,he2007mr,griswold2002generalized,archibald2016image}, ultrasound imaging \cite{wagner_compressed_2012,chernyakova_fourier-domain_2014} and synthetic radar imaging \cite{alonso_novel_2010,chen_time-frequency_2002}. The reconstruction methods in the literature can be roughly classified into two categories: the discrete models and the continuous ones. The discrete models view the underlying image as a discrete vector with certain \textit{fixed} resolution and usually obtain its approximation through solving a discrete optimization problem consisting of a fidelity term and a regularity term. There are various regularity terms used in the literature, such as total variation \cite{rudin1992nonlinear,osher2005iterative}, total generalized variation \cite{bredies2010total,knoll2011second}, and total fractional variation \cite{chen2013fractional,chen2013fractional2,zhang2015total} \etc. Moreover, many efficient algorithms such as alternating direction method of multipliers (ADMM), primal-dual methods have been proposed  \cite{boyd2011distributed,goldstein2009split,parikh2014proximal,zulfiquar_ali_bhotto_improved_2015,wahlberg_admm_2012,becker_nesta:_2011,osherfast,DBLP:journals/mpc/BeckerCG11} to solve corresponding optimization problems. On the other hand, the continuous models consider the underlying image as a piece-wise smooth function and recover the image from a function approximation point of view. One of its advantages is the flexibility in setting resolution and it has been successfully employed in the reconstruction of super-resolution images \cite{CPA:CPA21455,deng2016single}. It has also been shown to have superior performance in generalized/infinite-dimensional compressive sensing reconstruction \cite{adcock_generalized_2015,adcock2013breaking}.

Image reconstruction has usually been formulated as an optimization problem that minimizes an energy functional in the following form:
\begin{equation*}
\min \mathcal{L}(x) + \mathcal{R}(x),
\end{equation*}
where $\mathcal{L}(x)$ is a fidelity term depending on an empirical estimation of the distribution of noise, and $ \mathcal{R}(x)$ is a regularity term with a prior estimation of the structure of the underlying image. For instance, $\mathcal{L}(x)$ is often a least squares term while the noise is assumed to be Gaussian, and other formulations could also be found in \cite{li2015multiphase}. On the other hand, a widely used regularity term $\mathcal{R}(x)$ is the $l_1$ type constraint incorporating certain sparsity prior knowledge. Such sparsity may come from edge estimation \cite{lu2014edge,guo2012edge,cai2016image}, wavelet transformation \cite{dong_mra-based_2012,song2013approximating,zhang__2013,choi2017edge}, different orders of total variation  \cite{bredies2010total,zhang2015total,chan2000high,chumchob2011fourth} \etc.

Most of existing methods solve the above optimization problem with various fidelity terms and regularity terms in either continuous or discrete settings. We will refer to them as one stage methods in this paper. We will leverage both discrete and continuous models to develop a \textit{two-stage} Projection Correction Modeling (PCM), in which the first stage (P-stage) employs a continuous model and the second stage (C-stage) imposes a discrete regularization/penalty term on the model. In particular, for image reconstruction from Fourier measurements, we will later show in numerical experiments that the proposed two stage PCM has a superior performance comparing with other popular one stage methods.

We demonstrate the idea of PCM with a reconstruction problem. The given data are some finite uniform or non-uniform Fourier measurements. The goal is to reconstruct the underlying image from these measurements. At the P-stage, we consider the underlying image as a function $f$ in a processing domain that is usually a Hilbert space spanned by some basis such as polynomials or wavelets.  We will find an ``optimal'' approximation (projection) in the processing domain by minimizing a certain data fidelity term. In other words, the P-stage projects the Fourier (k-space) measurements into another processing domain that has an accurate representation of the underlying function. We also point out that the P-stage could also be viewed as a dimension reduction step, since we often choose a projection on a much lower dimension subspace. It will also help to reduce the computational time at the second stage. Due to the noise in the measurements and/or imperfect selection of the basis, the approximation in the P-stage will also contain errors. To further improve the reconstruction, we will impose a discrete regularization at the C-stage.

At the C-stage, we will find a ``corrected'' approximation in the same processing domain by minimizing the sum of a date fidelity term and a regularity term. The data fidelity is the difference between the corrected approximation and the approximated function obtained at the P-stage. On the other hand, we will employ a regularity term on the discrete vector that is the evaluation of the corrected approximation function on discrete grids. In particular, we will consider a hybrid regularity combining total variation and total fractional variation. The total variation (TV) regularization has very good performance in keeping edges in the reconstruction but suffers from the staircase artifact which causes oil-painted blocks. It is mainly due to the fact that TV is a local operator. On the other hand, the total fractional variation (TFV) is a recent proposed regularization term in image processing and has achieved promising results \cite{chen2013fractional,chen2013fractional2,zhang2015total,zhang2012class,verdu2009fractional,melbourne2012using}. Imposing TFV could reduce such artifact due to its non-local nature. However, the edges are damped in the reconstructed image based on TFV regularization. Therefore, we will adopt a hybrid regularity with TV on the ``edges'' and TFV on the ``smooth'' part.

We remark that edge detection becomes an important task in our method, since we will not know where the true edges are before we proceed. We would like to mention that there is another research pipeline in multi-modality image reconstruction where computed tomography (CT) and MRI scanning would be run at the same time and CT images are used to enhance the performance of MRI as well as reducing the processing time \cite{lu2014edge,brankov2002multi,cui2014total,james2014medical}. The set of edges will be given since the CT is much faster than MRI. However, we consider a more challenging task in this paper where the edges are not known and will be reconstructed recursively in our algorithm.

We briefly summarize our contributions below:
\begin{enumerate}[(1)]
	\item We propose a novel two-stage PCM method of image reconstruction that leverages advantages of both discrete and continuous models.
	\item We employ a precise and efficient edge based variational constraint in the regularity term, while the edge is determined through combining techniques of image morphology and \textit{thresholding} strategy.
	\item We develop an efficient proximal algorithm of solving the proposed model.
\end{enumerate}

The rest of the paper is organized as follows. We introduce the proposed PCM framework for image reconstruction from Fourier measurements in \Cref{PCM section}. We develop in \Cref{algorithms} an efficient proximal algorithm for the general PCM model and further employ it to derive an algorithm for a specific PCM-TV-TFV model. Numerical experiments and comparisons are presented in \Cref{numerical experiments}. We finally make some conclusion remarks in \Cref{conclusions}.

\section{Projection Correction Modeling}\label{PCM section}
We will present the proposed PCM framework for image reconstruction from Fourier measurements in this section. To this end, we first give a brief introduction of the image reconstruction problem.

Suppose the underlying image is a real-valued function $f: \Omega \to \R$, with $\Omega \subseteq \R^2$. We are given its Fourier data $\hat{f}$ in the following form:
\begin{equation}\label{eq:Fourierdata}
\hat{f} = \mathcal{S}\F f + \epsilon,
\end{equation}
\begin{itemize}
	\item[$\bullet$] $\mathcal{S}$ is the sampling operator (might be uniform or non-uniform), 
	\item[$\bullet$] $\F$ is the continuous Fourier transform as
	\[
	\F f(w) =  \int_{\bR^2} f(x)  \exp(-2\pi i w\cdot x) dx.
	\]
	\item[$\bullet$] $\epsilon \in \C^m$ is random noise.
\end{itemize}
Our goal is to recover the underlying image $f$ from the given Fourier data. 

The challenges come from the non-uniformness of $\mathcal{S}$ as well as the appearance of the noise $\epsilon$. It is well received that most images are piece-wise smooth with potential jumps around the edges. The inverse Fourier transform will not work directly here. The non-uniformness of the samples in the frequency domain (k-space) will make the inverse process unstable and might bring extra approximation errors when the sampled data contain noises. Moreover, the Fourier basis is amenable to Gibbs oscillations artifacts in representing a piece-wise smooth function. To overcome such challenges, we will propose a two-stage PCM framework below.  

We point out that the PCM framework can be viewed as a generalization of the classical multi-scale Galerkin method in finite element analysis. In contrast to the classical triangle or polyhedron mesh segmentation in a single scale, the region has elements in different scales. The advantage of doing so is to increase the numerical stability of the solver. More details can be found in chapter 13 of \cite{gockenbach2006understanding}.

\begin{figure}[tbhp]
\centering
\includegraphics[width=\columnwidth]{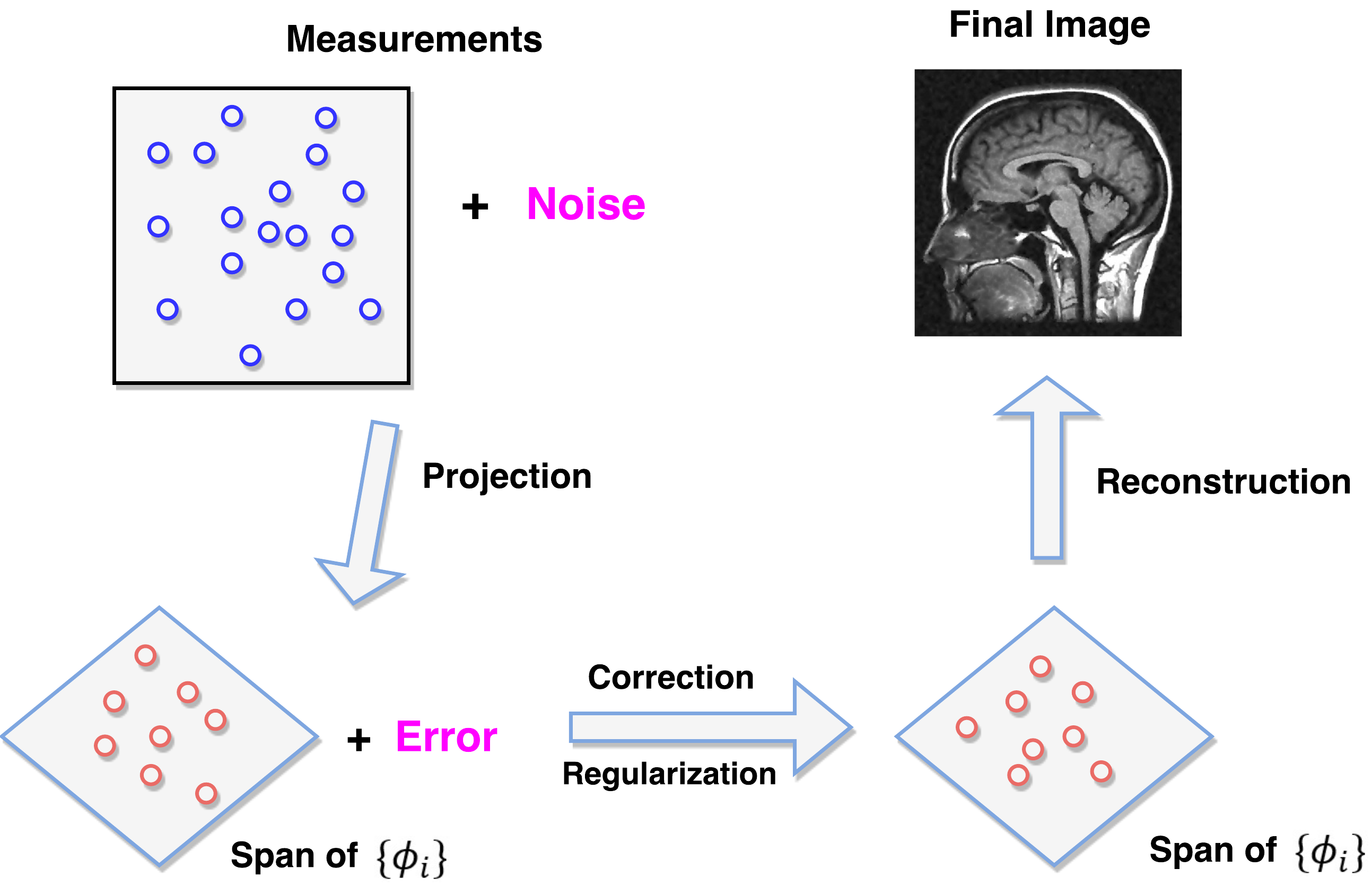}
\vskip 0.2in
\caption{General PCM framework}\label{fig_PCM_framework}
\end{figure}

We present the general idea of such a PCM framework in \Cref{fig_PCM_framework}. We shall next introduce the P-stage and the C-stage in PCM framework with more details.

\subsection{The P-stage}
 Suppose $\{\phi_1, \phi_2,...,\phi_n\}$ is a basis of a subspace of the processing domain and we will find an ``optimal'' approximation 
\begin{equation*}
\tilde{f}=\sum_{j=1}^n (\vc_f)_j \phi_j,
\end{equation*}
where its coefficients $\vc_f$ is obtained through solving the following least squares problem
\begin{equation}\label{eq:Pstage}
\vc_f = \argmin_\vc \| \mathcal{S}\mathcal{F}\Phi \vc  - \hat{f}\|_2^2,
\end{equation}
where $\Phi\vc=\sum_{j=1}^n c_j\phi_j$. The above least squares problem could be solved efficiently by conjugate gradient solvers.

The performance of the P-stage depends on the selection of the basis $\{\phi_i\}_{i = 1}^{n}$ according to the prior knowledge of the property of the image. We point out that the above framework has also been used in \cite{song2013approximating, gelb_frame_2014} for approximating the inverse frame operator with admissible frames. Analysis of the approximation error has also been discussed there. In particular, Fourier basis has been used to obtain a stable and efficient numerical approximation of a smooth function from its non-uniform Fourier measurements. However, Fourier basis might not be appropriate to represent a piece-wise smooth function. Instead, we will consider wavelets including classical ones such as Haar, Daubechies wavelet or more recent ones such as curvelet \cite{starck2002curvelet,ma2010curvelet}, shearlet \cite{easley2008sparse,guo2014new}  \etc, which have a more accurate representation for piece-wise smooth functions. In this regard, it could also viewed as a generalization of the admissible frame method in \cite{song2013approximating, gelb_frame_2014}. In this paper, 1D and 2D Haar wavelets are used in the numerical experiments, but the general PCM framework could also work with other wavelets.

We remark that even with a reasonable selection of basis $\{\phi_i\}_{i = 1}^n$, $\vc_f$ from \cref{eq:Pstage} might not be accurate due to the noise in the data. We proceed to improve it in the correction stage to alleviate those effects.

\subsection{The C-stage}
We will find a ``corrected'' approximation 
\begin{equation*}
g=\sum_{j=1}^n (\vc_g)_j \phi_j,
\end{equation*}
in the same processing domain through solving the following regularization optimization problem
\begin{equation*}
\min_{\vc_g} \frac{1}{2}\|\vc_g - \vc_f\|^2 + \mathcal{R}(\vc_g),
\end{equation*}
where $\mathcal{R}(c_g)$ imposes prior knowledge based on the understanding of the processing domain. The idea is to find a new set of coefficients that are close to $\vc_f$ found in P-stage such that the new approximation satisfies some regularities. In terms of optimization, it provides the proximal guidance during the algorithm implementation. 

\subsubsection{TV-TFV Regularity}
We next discuss the choice of the regularity term $\cR(\vc_g)$. One popular choice in the literature uses $\|c_g\|_1$ by assuming the underlying function has a sparse representation in the processing domain. However, it might not be the best choice for the problem of image reconstruction from Fourier measurements. It does not incorporate the property of images that are piece-wise smooth and contain many textures/features. Moreover, it might have the bias issue in statistics literature that involves the modeling error brought by imperfect selection of basis  $\{\phi_i\}_{i = 1}^n$. That is, a regularity term on the coefficients $\|c_g\|_1$ might not be the best choice of alleviating the bias issue. Instead, we would impose regularization in image/function values on a discrete grid through some prior knowledge about the underlying images such as the piece-wise smoothness, textures/features.  In particular, we shall employ a regularity that combines total variation (TV) and total fractional variation (TFV). To this end, we first review some definitions and notations of fractional order derivatives.

We point out that there are several definitions of fractional order derivatives, such as Riemann-Liouville (RL), Grunwald-Letnikov (GL), Caputo  etc. We will employ the RL definition \cite{miller1993introduction,gorenflo1997fractional} in this paper. The left, right and central RL derivatives of order $\alpha \in (n-1, n)$, $n \in \mathbb{N}$ for a function $f(x)$ supported on an interval $ [a,b] $,  are defined by
\[
_aD_x^\alpha f(x) = \frac{1}{\Gamma(n-\alpha)}\frac{d^n}{dx^n}\int_{a}^{x} (x-\tau)^{n - \alpha -1}f(\tau)d\tau,
\]
\[
_xD_b^\alpha f(x) = \frac{(-1)^n}{\Gamma(n-\alpha)}\frac{d^n}{dx^n}\int_{x}^{b} (\tau-x)^{n - \alpha -1}f(\tau)d\tau,
\]
and 
\[
_aD_b^\alpha f(x) = \frac{1}{2} (_aD_x^\alpha f(x) + \text{ }_xD_b^\alpha f(x) \cdot (-1)^n),
\]
where $\Gamma(\alpha)$ is the Euler's Gamma function
\[
\Gamma(\alpha) = \int_{0}^{\infty} t^{\alpha - 1}e^{-t}dt.
\]

We next introduce the discretization of the RL fractional derivative. In the simplicity of presentation, we will show the 1D discretization over an interval $[a,b]$. The discretization over a 2D regular domain will be a direct generalization of this procedure along horizontal and vertical directions. We consider the following $n$ equidistant nodes on $[a,b]$:
$$x_i = \frac{(i - 1)(b-a)}{n} + a, \hspace{0.2in} i = 1,2, ..., n.$$ 
Let $L^{(\alpha)}$ and $R^{(\alpha)}$ be the matrix approximations of the left and right-sided Riemann-Liouville $\alpha-$order derivative operator $_aD_x^\alpha$ and $_xD_b^\alpha$ accordingly. With Dirichlet boundary conditions that $f(a) = f(b) = 0$,  it follows \cite{podlubny2000matrix} that $L^{(\alpha)}$ and $R^{(\alpha)}$ are two triangular strip matrices with the following structure:
\[
L^{(\alpha)}_n = n\left(\begin{array}{cccccc}
w_0^\alpha & w_1^\alpha & \cdots & \cdots & w_{n-1}^\alpha & w_{n}^\alpha \\ 
0 & w_0^\alpha & w_1^\alpha & \ddots & \ddots & w_{n-1}^\alpha \\ 
0 & 0 & \ddots & \ddots & \ddots & \vdots \\ 
\vdots & \ddots & \ddots & w_0^\alpha & w_1^\alpha & \vdots \\ 
0 & \ddots & \ddots & 0 & w_0^\alpha & w_1^\alpha \\ 
0 & 0 & \cdots & 0 & 0 & w_0^\alpha
\end{array}  \right) 
\]
and
\[
R^{(\alpha)}_n = n\left(\begin{array}{cccccc}
w_0^\alpha & 0 & 0 & \cdots & 0 & 0 \\ 
w_1^\alpha & w_0^\alpha & 0 & \ddots & \ddots & 0 \\ 
\vdots & \ddots & \ddots & \ddots & \ddots & \vdots \\ 
\vdots & \ddots & \ddots & w_0^\alpha & \ddots & 0 \\ 
w_{n-1}^\alpha & \ddots & \ddots & w_1^\alpha & w_0^\alpha & 0 \\ 
w_n^\alpha & w_{n-1}^{\alpha} & \cdots & \cdots & w_1^\alpha & w_0^\alpha
\end{array}  \right) ,
\]
where $w_0^\alpha = 1$, $w_j^\alpha = (-1)^j\binom{\alpha}{j}$. These coefficients then can be constructed iteratively:
\[
w_j^\alpha = (1 - \frac{1+ \alpha}{j})w_{j-1}^\alpha,\hspace{0.2in} j = 1,2,...,n.
\]
Furthermore, when $\alpha \in (1,2)$, the matrix approximation of the central RL derivative $C^{(\alpha)}$ will then become
\[
C^{(\alpha)}_n = \frac{1}{2}(L^{(\alpha)} + R^{(\alpha)}).
\] 

Let $U \in \R^{N}$ be the given discretized image $u$ under an ordinary xy-coordinate with pixel values lexicographically ordered in a column vector. For simplicity, we assume that the original image $u$ is in square size with $n = \sqrt{N}$ rows and columns. Let $\otimes$ denote the Kronecker product. It follows from applying the central RL derivative $C^{(\alpha)}$ to the images along the $x$-direction that
\[
u_x^{(\alpha)} = (I_n\otimes C^{(\alpha)}_n)U,
\] where $I_n \in \R^{n\times n}$ is the identity matrix. Similarly, along the $y-$direction we will have
\[
u_y^{(\alpha)} = ( C^{(\alpha)}_n \otimes I_n)U.
\]
The procedure will be the same for other choices of $L_n^{(\alpha)}$ and $R_n^{(\alpha)}$. The $l_1$ norm regularization over $u_x^{(\alpha)}$ and $u_y^{(\alpha)}$ leads to the total fractional variation models \cite{zhang2015total}. 

We remark that even though $C_n^{(\alpha)}$ is a dense matrix, the matrix  $I_n\otimes C^{(\alpha)}_n$ will be sparse. Furthermore, we observe that $w_j^{\alpha}$ decays very fast. For example, take $\alpha = 1.3$, the first few $w_j$'s
\begin{align*}
&w_0 = 1, \hspace{0.1in} w_1 = -1.3, \hspace{0.1in} w_2 = 0.195, \hspace{0.1in} w_3 = 0.0455, \hspace{0.1in} \\
& w_4 = 0.0493, \hspace{0.1in}w_5 = 0.01,\hspace{0.1in} w_6 = 0.006  ...
\end{align*}
To further enhance the sparsity of the operation matrix, one can truncate $w_j$'s at a certain level to improve the efficiency of the program.

We are now ready to introduce the C-stage in the following form:
\begin{align}
	\hspace{0.2in}&\min_{\vc_g} \frac{1}{2}\|\vc_g - \vc_f\|^2 + \mu_t \|\nabla g|_{\Gamma}\|_1 + \mu_f \|\nabla^\alpha g|_{\Gamma^c}\|_1 \label{eq:Cstage} \\
	& s.t. \hspace{0.1in} g = \Phi \vc_g.
\end{align}
where $\Gamma$ is an open domain centered around the edge part of the image, $\nabla^\alpha$ is the discretized $\alpha$-order fractional differential operator.

We point out that an important question in the above model \cref{eq:Cstage} is to select the  $\Gamma$ set.  It denotes an open region centered around the edges rather than the edges themselves. The reason is that the set of edges has zero measure in $\R^2$ and it will not be very meaningful to consider the 2D total (fractional) variation on that.

\subsubsection{Construction of the $\Gamma$ set }
We shall next discuss the construction of the $\Gamma$ set.  We would first detect the edges of the image reconstructed from P-stage.  In order to obtain accurate edges, our approach is to find a rough estimation at the initial step, and then update it iteratively with the reconstructed image during the implementation of the reconstruction algorithm. Specifically, we would use the first few iterations of the total variation model \cite{rudin1992nonlinear} as a warm start to obtain an initial estimate of the edges. Meanwhile, we will take an early termination in such an iterative method to avoid staircase artifacts. We would then apply these filter edge detectors such as Sobel, Canny and Prewitt filters \cite{canny1986computational,szeliski2010computer,davies2012computer} on the initial result to improve the accuracy of edges.

\begin{figure}[tbhp]
	\centering
	$\vcenter{\hbox{\subfloat[Original Image]{\includegraphics[width=.3\linewidth]{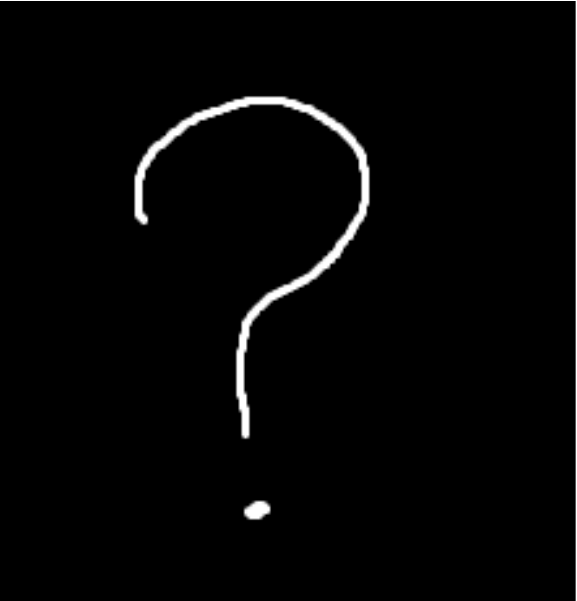}}}}\hspace{0.3in}
	\vpointer \hspace{0.3in}
	\vcenter{\hbox{\subfloat[Dilated Image]{\includegraphics[width=.3\linewidth]{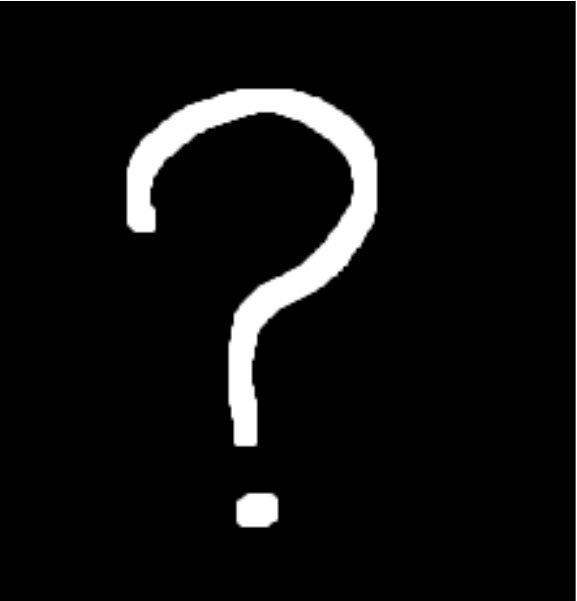}}}}$
	\caption{Illustration of Image Dilation.}\label{fig:dialation}
\end{figure}

Once the edges are obtained, an initial $\Gamma$ set can be easily constructed via morphologic dilation \cite{serra1982image}, see \Cref{fig:dialation} for a visual illustration. 

We would then update the $\Gamma$ set through a few iterations of the previous results. We next discuss the update of the $\Gamma$ set in more details. We denote the $\Gamma$ set obtained at the $i$th iteration by $\Gamma_i$. When estimating $\Gamma_k$, we would use the previous $\Gamma_i$'s:
\begin{equation*}
\Gamma_k = \texttt{round}\bigl(\frac{1}{k-1}\sum_{i = 1}^{k-1} \Gamma_i \bigr),
\end{equation*}
where $\texttt{round}(\cdot)$ is the standard round function that returns the closest integer. This is to ensure that $\Gamma_k$ will be a binary matrix. A more general thresholding scheme with a given distribution $\mathcal{D}$ at a certain confidence level $t \in [0,1)$ will be
\begin{equation*}
\Gamma_k = \Bigg({\sum_{i = q}^{k - 1} w_i \Gamma_i >t}\Bigg),
\end{equation*}
for some $1 \leq q \leq k-1, q \in N_+$, $w_i \sim \mathcal{D}$, $w_q\leq w_{q+1} \leq \cdots \leq w_{k-1}$, $\sum_{i = q}^{k-1}w_i = 1$.

We could further reduce the computational cost by thresholding on the number of iterations. We point out that the reconstructed image will become more accurate during the iterations of the algorithm and there will be little variance in the detected edges after certain iterations. That is, it is a reasonable to use a small number of iterations in edge detection. 

We summarize the procedure for reconstructing the initial $\Gamma$ set as follows:
	\begin{enumerate}
		\item Warm up and edge detection. 
		\begin{enumerate}
			\item Run TV model for a few iterations (in our experiment $3\sim 5$ will be enough).  This depends on a rough estimate of the noise level. A quick noise variance estimation can be found at \cite{rank1999estimation}.
			\item Implement filer based edge detector such as Canny, Sobel filters to detect the edge set.
		\end{enumerate}
		\item Obtain $\Gamma$ through image dilation.
	\end{enumerate}
The $\Gamma$ set is updated as the underlying image is iteratively updated.

\subsection{PCM-TV-TFV in Image Denoising}
We shall summarize the P-stage and the C-stage to present the complete process of the two stage PCM-TV-TFV model as follows:
		\begin{align}\label{eq:PCM-TV-TFV}
		\begin{split}
		\textbf{P}:\hspace{0.2in} & \vc_f = \argmin_\vc \|\mathcal{S}\mathcal{F}\Phi \vc  - \hat{f}\|_2^2, \\
	\textbf{C}:	\hspace{0.2in}&\min_{\vc_g} \frac{1}{2}\|\vc_g - \vc_f\|^2 + \mu_t \|\nabla g|_{\Gamma}\|_1 + \mu_f \|\nabla^\alpha g|_{\Gamma^c}\|_1 \\
		& s.t. \hspace{0.1in} g = \Phi \vc_g.
		\end{split}
		\end{align}
with $\Gamma$ is an open domain centered around the edges.

We will demonstrate the advantages of the above proposed two stage model through numerical comparisons with other popular models later.

\section{Algorithms}\label{algorithms}
In this section we shall develop a proximal algorithm scheme for solving the general PCM optimization problem. Moreover, we will also introduce a split Bregman scheme for solving the TV-TFV regularity problem. We would then combine them to derive a specific algorithm for the PCM-TV-TFV model.

\subsection{General PCM Model Solver} 
We shall first consider the following general PCM model with a general regularity term $\cR$:
\begin{align*}
\textbf{P}:\hspace{0.2in} & c_f = \argmin_c \|\mathcal{S}\mathcal{F}\Phi \vc  - \hat{f}\|_2^2 \\
\textbf{C}:\hspace{0.2in} & \min_{\vc_g} \frac{1}{2}\|\vc_g - \vc_f\|_2^2 + \mathcal{R}(\vc_g).
\end{align*}

We will introduce a general proximal algorithm for solving the above PCM optimization problem. We remark that the proximal algorithms refer to a class of algorithms that are widely used in modern convex optimization literature,  for a comprehensive survey, see \cite{parikh2014proximal}. To this end, we first review the definition of the proximal operator : for a convex function $\mathcal{R}$, the proximal operator $\mathbf{prox}$ \cite{parikh2014proximal} is defined as
\[
\mathbf{prox}_{\lambda \mathcal{R}}(\vv) = \argmin_{\vc} \biggl\{\mathcal{R}(\vc) + \frac{1}{2\lambda}\|\vc - \vv\|_2^2 \biggr\}.
\]
We point out that computing the proximal operator is equivalent to solve a trust region problem \cite{dennis1996numerical,parikh2014proximal}. Moreover, we could derive closed forms of such proximal operators for many popular functions. For example, if $\mathcal{R}(\vc) = \vc^T A \vc$ for some symmetric positive semidefinite matrix $A$, then 
\[
\mathbf{prox}_{\lambda \mathcal{R}}(\vv) = (A + 1/\lambda I)^{-1}\vv/\lambda.
\] 
If $\mathcal{R}(\vc) = \|\vc\|_1$, we have 
\[
(\mathbf{prox}_{\lambda \mathcal{R}}(\vv))_i = \begin{cases}
v_i - \lambda & \mbox{ if } v_i > \lambda, \\
v_i + \lambda & \mbox{ if } v_i < -\lambda, \\
0 & |v_i| \leq \lambda.
\end{cases} 
\]
which is the component-wise soft thresholding operator. 

Consequently, we present an efficient proximal algorithm in \cref{PCM_algorithm1} for solving the general PCM optimization problem, based on  Nesterov's accelerated gradient method as well as the FISTA algorithm \cite{nesterov1983method,beck2009fast,boyd2011distributed}. 

\begin{algorithm}[htbp]
	\caption{Accelerated Proximal Algorithm}
	\label{PCM_algorithm1}
	\begin{spacing}{1.1}
	\begin{algorithmic}[1]
		\REQUIRE{$\hat{f}$, $\Phi$, $\vc_g^0$, $\vd^0$, $\lambda^0$}
		\ENSURE{Reconstructed image $g$}
		\STATE \textbf{P-stage:} Construct $\mathcal{S}\mathcal{F}\Phi$ from (2), solve $\vc_f = (\mathcal{S}\mathcal{F}\Phi)^\dagger \hat{f}$.
		\STATE \textbf{C-stage:} \WHILE{not converged}  
		\STATE $\vc_g^k = \mathbf{prox}_{\lambda^k\mathcal{R}} (\vd^{k} - \lambda^k(\vd^k - \vc_f));$
		\STATE $\vd^{k+1} = \vc_g^{k} + \lambda_k(\vc_g^k - \vc_g^{k-1});$
		\STATE $\lambda^k = \frac{k}{k+3};$
		\ENDWHILE
		\STATE \textbf{Construct} $g = \Phi \vc_g^k$.
	\end{algorithmic}
\end{spacing}
\end{algorithm}

We observe that \cref{PCM_algorithm1} involves a linear program and an iterative proximal operator evaluation problem. Both of them can be computed very efficiently. The update step for $\lambda^k$ is taken from the FISTA algorithm \cite{nesterov1983method,beck2009fast,boyd2011distributed}, which will accelerate the convergence.

\subsection{TV-TFV Regularity Solver}
\begin{figure}[htbp]
	\centering
	\includegraphics[width=1\textwidth,height=6.6cm]{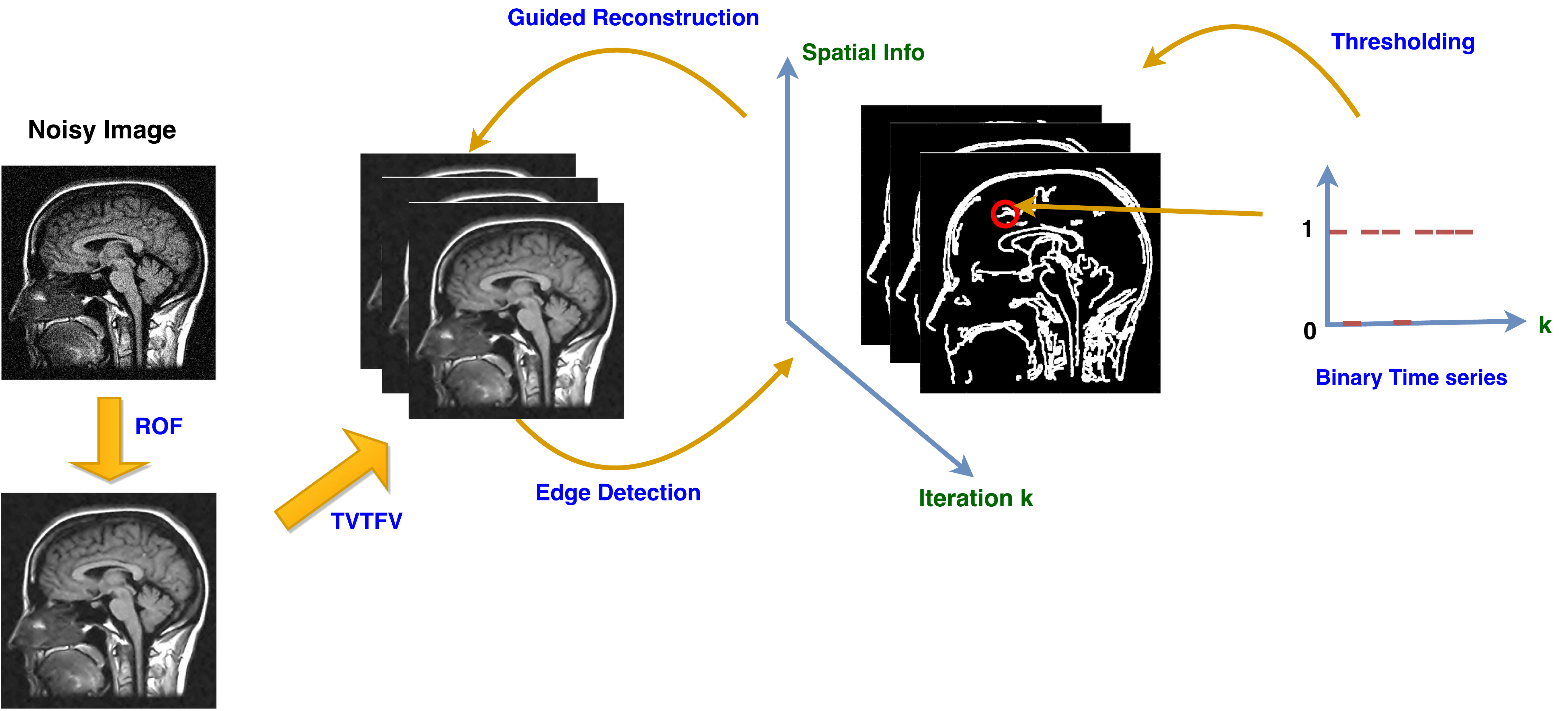}
	\caption{Illustration of the TV-TFV regularization scheme in image denoising problem. }\label{fig:TVTFVworkflow}
\end{figure}

In this subsection we shall consider the following TV-TFV regularity problem:
\begin{equation}\label{eq:TV-TFVdenoising}
\min_{f} \frac{1}{2}\|f - \f\|^2 + \mu_t \|\nabla f|_{\Gamma}\|_1 + \mu_f \|\nabla^\alpha f|_{\Gamma^c}\|_1,
\end{equation}
	where $\Gamma$ again is an open domain centered around the ``edge'' set.  If $\mu_f$ is set to be 0 and the width of $\Gamma$ is large enough, the above model will reduce to the classical TV denoising model.
	\[
	\min_{f} \frac{1}{2}\|f - \f\|^2 + \mu_t \|\nabla f\|_1.
	\]
	 Similarly, if $\Gamma$ is set to be $\emptyset$, it will become the total fractional variational model, 
	 \[
	 \min_{f} \frac{1}{2}\|f - \f\|^2  + \mu_f \|\nabla^\alpha f\|_1.
	 \]
We remark that this model is used in the image denoising problem. The work-flow of image denoising problem with TV-TFV regularization is summarized in \Cref{fig:TVTFVworkflow}. 

We shall introduce an algorithm of solving the above TV-TFV regularization problem. We will start with a warm up procedure to get an estimate of the edge regions. We would then employ the split Bregman method \cite{goldstein2009split,yin2008bregman} to derive the algorithm.  We present it in \cref{TV-TFV_Denoising}, where $\textbf{shrink}$ is the soft shrinkage function defined by
\[
\textbf{shrink}(x, \alpha) = \begin{cases}
x - \alpha & \mbox{ if } x > \alpha, \\
x + \alpha & \mbox{ if } x < -\alpha, \\
0 & |x| \leq \alpha.
\end{cases}
\]

\begin{algorithm}[htbp]
		\caption{TV-TFV Denoising Algorithm}
		\label{TV-TFV_Denoising}
		\begin{algorithmic}[2]
			\STATE \textbf{Initialize} $\bar{\Gamma}_0 = \Omega$, $f^1 = \f$,   confidence level t, length of stored sequence $n$. 
			\STATE \textbf{Warm up} with $3\sim5$ TV iterations and update $\bar{\Gamma}_0$ to $\bar{\Gamma}_1$. 
			\FOR {$k = 1,2,...$ }
			\IF {$k > n$} 
			\STATE {$\bar{\Gamma}_k = (\sum_{i = k-n+1}^{k}\bar{\Gamma}_i)>n*t$; }
			\ENDIF
			\STATE \textbf{Update} $\Gamma_k$ = dilate($\bar{\Gamma}_k$);
			\STATE \textbf{Update} $d^k = \nabla f|_{\Gamma_k}$, $e^k = \nabla^\alpha f|_{\Gamma_k^c}$.
			\STATE \textbf{Solve} f-subproblem 			
			\begin{align*}
			f^{k+1} = \argmin_f &\frac{1}{2}\|f-\hat{f}\|_2^2 + \mu_t\|\nabla f|_{\Gamma_k} - d^k + dd^k\|_2^2 \\+ &\mu_f\|\nabla^\alpha f|_{\Gamma_k} - e^k + ee^k\|_2^2 
			\end{align*}  \vskip -0.1in	
			\STATE	$d-$subproblem 	$$d^{k+1} = \textbf{shrink}(\nabla f^{k+1}|_{\Gamma_k}+dd^k, \mu_t/\lambda_t)$$ \vskip -0.1in
			\STATE	$dd-$subproblem $$dd^{k+1} = dd^k + \gamma_1(\nabla f^{k+1}|_{\Gamma_k}-d^{k+1})$$ \vskip -0.1in
			\STATE $e-$subproblem 
			$$e^{k+1} = \textbf{shrink}(\nabla^\alpha f|_{\Gamma_k} + ee^k, \mu_f/\lambda_f)$$ \vskip -0.1in
			\STATE $ee-$subproblem
			$$ee^{k+1} = ee^k + \gamma_2(\nabla^\alpha f|_{\Gamma_k} - e^{k+1})$$  \vskip -0.1in
			\IF{ converged }
			\STATE \textbf{break}
			\ENDIF
			\ENDFOR
		\end{algorithmic}
\end{algorithm}

We will demonstrate the proposed TV-TFV regularization at the C-stage has a superior performance by numerical comparisons with the TV denoising model and the TFV denoising model in \Cref{numerical experiments}.

\subsection{PCM-TV-TFV Solver}
We now present an algorithm of solving the PCM-TV-TFV model \cref{eq:PCM-TV-TFV} for image reconstruction in \Cref{PCM-TV-TFV Algorithm}. It follows from a direct application of ADMM \cite{boyd2011distributed} and an accelerated ADMM algorithm \cite{goldstein2014fast}. 

\begin{algorithm}[htbp]
	\caption{PCM-TV-TFV Algorithm}
	\label{PCM-TV-TFV Algorithm}
	\begin{algorithmic}[2]
		\REQUIRE{$\hat{f}$, $\Phi$, $\vd^0$, $\lambda^0$}
		\ENSURE{Reconstructed image $g$}
		\STATE \textbf{P-stage:} Construct $\mathcal{S}\mathcal{F}\Phi$ from (2), solve $\vc_f = (\mathcal{S}\mathcal{F}\Phi)^\dagger \hat{f}$.
		\STATE \textbf{C-stage:} \textbf{Initialize} $\bar{\Gamma}_0 = \Omega$, confidence level t, length of stored sequence $n$. 
		\STATE \textbf{Warm up} with $3\sim5$ TV iterations and update $\bar{\Gamma}_0$ to $\bar{\Gamma}_1$. 
		\FOR {$k = 1,2,...$ }
		\IF {$k > n$} 
		\STATE {$\bar{\Gamma}_k = (\sum_{i = k-n+1}^{k}\bar{\Gamma}_i)>n*t$; }
		\ENDIF
		\STATE \textbf{Update} $\Gamma_k$ = dilate($\bar{\Gamma}_k$);
		\STATE \textbf{Update} $d^k = \nabla (\Phi \vc_g^k)|_{\Gamma_k}$, $e^k = \nabla^\alpha (\Phi \vc_g^k)|_{\Gamma_k^c}$.
		\STATE \textbf{Solve} $\vc_g$-subproblem 			
		\begin{align*}
		\vc_g^{k+1} = \argmin_{\vc_g} &\frac{1}{2}\|\vc_g-\vc_f\|_2^2 + \mu_t\|\nabla (\Phi \vc_g)|_{\Gamma_k} - d^k + dd^k\|_2^2 \\+ &\mu_f\|\nabla^\alpha (\Phi \vc_g)|_{\Gamma_k} - e^k + ee^k\|_2^2 
		\end{align*}  \vskip -0.1in	
		\STATE	$d-$subproblem 	$$d^{k+1} = \textbf{shrink}(\nabla (\Phi \vc_g^{k+1})|_{\Gamma_k}+dd^k, \mu_t/\lambda_t)$$ \vskip -0.1in
		\STATE	$dd-$subproblem $$dd^{k+1} = dd^k + \gamma_1(\nabla (\Phi \vc_g^{k+1})|_{\Gamma_k}-d^{k+1})$$ \vskip -0.1in
		\STATE $e-$subproblem 
		$$e^{k+1} = \textbf{shrink}(\nabla^\alpha (\Phi \vc_g^{k+1})|_{\Gamma_k} + ee^k, \mu_f/\lambda_f)$$ \vskip -0.1in
		\STATE $ee-$subproblem
		$$ee^{k+1} = ee^k + \gamma_2(\nabla^\alpha (\Phi \vc_g^{k+1})|_{\Gamma_k} - e^{k+1})$$  \vskip -0.1in
		\IF{ converged }
		\STATE \textbf{break}
		\ENDIF
		\ENDFOR
	\end{algorithmic}
\end{algorithm}

\section{Numerical Experiments} \label{numerical experiments}
In this section we  demonstrate the superior performance of our proposed two stage PCM framework for image reconstruction from Fourier measurements. In particular, we will use numerical experiments to show that 
\begin{enumerate}[(1)]
\item The projection step itself can achieve accurate recovery in the case without noise and bias error;

\item The projection-correction with TV regularity (PCM-TV) has a better performance than many of the state-of-the-art continuous models;

\item The TV-TFV regularity leads to better results in image denoising;

\item The projection-correction with TV-TFV regularity (PCM-TV-TFV) model further improve the results of the PCM-TV.
\end{enumerate}

\begin{figure}[htbp]
			\centerline{
				\includegraphics[width=0.5\textwidth]{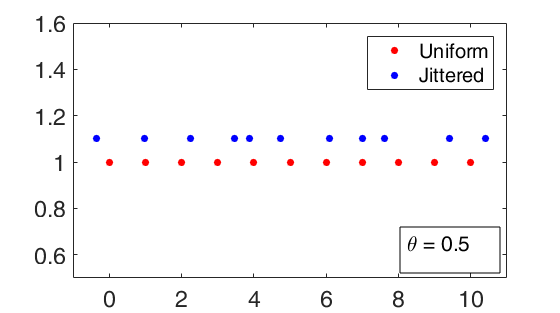}
			}
		\caption{Illustration of jittered and uniform sampling schemes.}\label{jitter_sampling}
		\end{figure}

We will focus on the function reconstruction from non-uniform measurements. In particular, we consider the jittered sampling in the frequency domain. That is, we assume the sampling is taken at the following frequencies:
\begin{equation*}
w_k = k + \eta_k, \hspace{0.1in} \text{} \eta_k \sim U[-\theta,\theta], k=-\frac{m}{2},...,\frac{m}{2}-1.
\end{equation*}
We display an example of the jittered sampling in \cref{jitter_sampling}.

\subsection{Accurate Recovery of the P-Stage}
We will use a simple example to demonstrate the accurate recovery of the P-stage when there is no noise and no bias error in the model. For a selected processing domain $\Hil_n = \tspan\{\phi_i\}_{i=1}^{n}$, we say it has no bias error  if the underlying function $f \in \Hil_n$.

We will use the classical Haar wavelets in the processing domain. In the 1D case, the mother wavelet $\psi(x)$ is
\begin{align*}
\psi(x) = \begin{cases}
1 & \mbox{if } 0\leq x < 1/2, \\
-1 & \mbox{if } 1/2\leq x <1, \\
0& \mbox{otherwise.}  \end{cases}
\end{align*}
and its descendants are $\psi_{n,k} = 2^{n/2}\psi(2^nx-k)$, $k = \mathbb{Z}, x\in [0,1)$. The 2D Haar wavelet is formulated by simple cross-product.

We consider the following piece-wise constant test function, 
\begin{align*}
f(x) = \begin{cases} -1/2 &\mbox{if } -1/8 \leq x < 1/4, \\ 
1 & \mbox{if } 1/4\leq x < 1/2, \\
-1 & \mbox{if } 1/2\leq x <5/8, \\
1/2 & \mbox{if } 5/8 \leq x < 3/4, \\
0 & \mbox{otherwise.}  \end{cases}
\end{align*}
  
\begin{figure}[htbp]
\centering
\subfloat[Projection]{\includegraphics[width=\globalwidth\columnwidth]{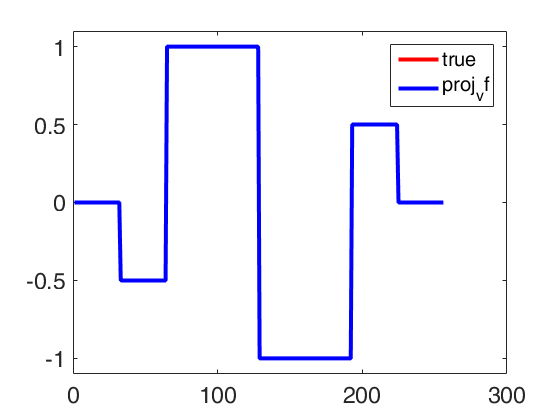}}
\subfloat[Truncated Fourier]{\includegraphics[width=\globalwidth\columnwidth]{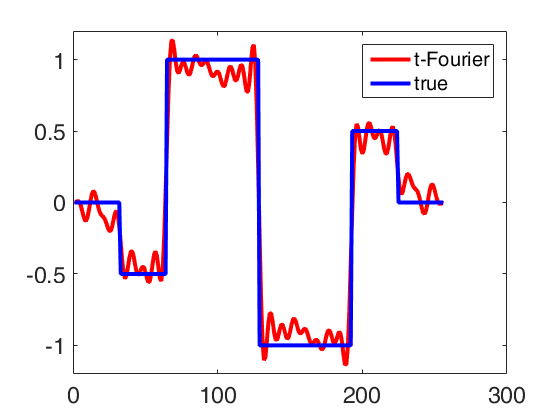}}\\
\centering
\subfloat[Proposed]{\includegraphics[width=\globalwidth\columnwidth]{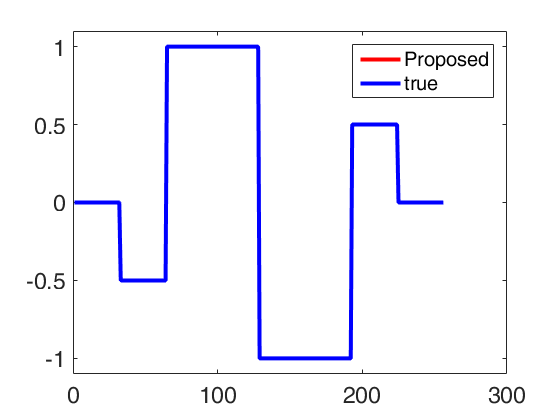}}
\subfloat[Difference]{\includegraphics[width=\globalwidth\columnwidth]{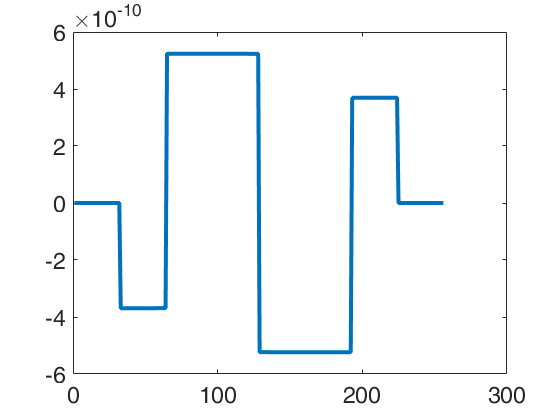}}
\caption{	Bias free reconstruction for piece-wise constant signal without noise: a) Ground truth signal and its projection on $\Hil_n$. b) Reconstruction from truncated Fourier series which suffers from Gibbs oscillation. c) Reconstruction from projection stage.  Here $m = 128$, $n = 32$, $resol = 1/256$, $\theta = 0.25$. d) The difference of the reconstruction and the ground truth.} \label{Bias-free}
\end{figure}

One can verify that the support of this test function is a subset of that of the selected Haar wavelets.  \Cref{Bias-free} shows that our proposed projection stage achieves a very accurate recovery for noiseless case.

\subsection{PCM-TV Model}
\begin{figure}[htbp]
	\centerline{			
		\subfloat[]{\includegraphics[width=\globalwidth\columnwidth]{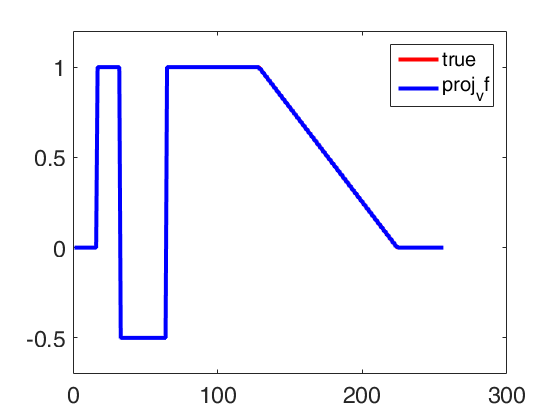}}
		\subfloat[]{\includegraphics[width=\globalwidth\columnwidth]{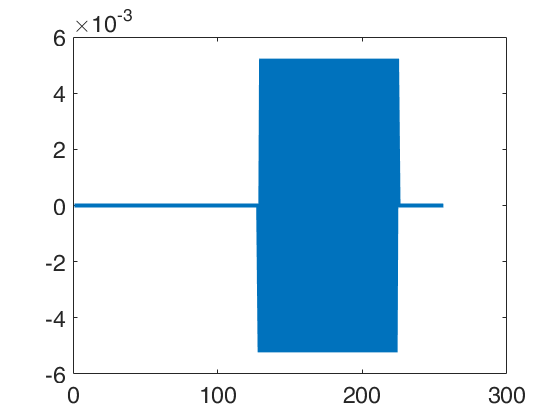}}
	}
	\caption{(a): Piece-wise linear function (red) and its projection onto $\Hil_{n}$, $n = 128$ (blue).  (b): the difference between the two.} \label{bias_signal}
\end{figure}

We consider in this subsection the case where the reconstruction contains bias error. In other words, the underlying function does not lie in the finite dimensional subspace spanned by the chosen basis. We will consider the following piece-wise linear test function $f(x)$:
\begin{align*}
f(x) = \begin{cases}
1 & \mbox{if } 1/16\leq x < 1/8, \\
-1/2 & \mbox{if } 1/8\leq x <1/4, \\
1 & \mbox{if } 1/4 \leq x < 1/2, \\
-\frac{8}{3}x + \frac{7}{3} & \mbox{if } 1/2 \leq x < 7/8, \\
0& \mbox{otherwise.}  \end{cases}
\end{align*}
We will use the same Haar wavelets to construct the processing domain. \Cref{bias_signal} displays the bias error.

Suppose we are given $m=256$ non-uniform Fourier measurements with some added Gaussian noise $\epsilon$ at various noise levels $\sigma = {\|\epsilon\|_2}/{\|\hat{f}\|_\infty}$. 

We will use TV as the regularity at the C-stage of our two stage PCM method and call it PCM-TV. Moreover, we will compare the proposed two stage PCM-TV with the following popular one stage method with different regularities:
	\begin{itemize}
		\item $\ell^1$ regularization model
		\begin{equation*}
\min_\vc \frac{1}{2} \|\mathcal{S}\mathcal{F}\Phi \vc - \hat{f}\|_2^2 +　\lambda \| \vc\|_1.
		\end{equation*}
		\item Tikhonov regularization model,
		\[
		\min_\vc \frac{1}{2} \|\mathcal{S}\mathcal{F}\Phi \vc - \hat{f}\|_2^2 +　\lambda \| \vc\|_2^2.
		\]
		\item Single stage TV (SS-TV) model, 
		\[
		\min_\vc \frac{1}{2} \|\mathcal{S}\mathcal{F}\Phi \vc - \hat{f}\|_2^2 +　\lambda \|\nabla \Phi \vc\|_1.
		\]
	\end{itemize}		
In particular, we will compare them in terms of a few different performance measurements including structural similarity (ssim) \cite{wang2004image},  peak signal-to-noise ratio (psnr), signal-to-noise ratio (snr) and relative error (rela\_err) defined as follows:
	\begin{align*}
		psnr &= 10\log_{10}\frac{d_1d_2(\max_{i,j}u_{ij})^2}{\|u - \hat{u}\|_F^2}, \\
		snr &= 10\log_{10} \frac{\|mean(\hat{u}) - u\|_F^2}{\|\hat{u} - u \|_F^2}, \\ rela\_err &= \frac{\|u - \hat{u}\|_2}{\|u\|_2},
	\end{align*}
	where $u \in \R^{d_1\times d_2}$ is the true image and $\hat{u}\in \R^{d_1\times d_2}$ is the reconstruction. We present the numerical results in \cref{table_all}. 
	\begin{table}[htbp]
		\captionsetup{position=top}
		\caption{Numerical comparison of the different models on different noise level. } \label{table_all}	
		\centering	
		\begin{tabular}{|l|l|l|l|l|l|}
			\hline
			$\sigma$ & Model    & ssim   & psnr  & snr   & rela\_err \\  \hlineB{3}
			0.1      & Proposed & \textbf{0.9283} & \textbf{39.14} & \textbf{33.58} & \textbf{0.0161 }     \\ \hline
			& $\ell^1$ regularization      & 0.8478 & 36.45 & 30.88 & 0.0220     \\ \hline
			& SS-TV    & 0.8466 & 35.72 & 30.16 & 0.0239      \\ \hline
			& Tikhonov & 0.7640 & 32.17 & 26.60 & 0.0360     \\ \hlineB{3}
			0.4      & Proposed & \textbf{0.8505} & \textbf{29.41} & \textbf{23.84} & \textbf{0.0495}     \\ \hline
			& $\ell^1$ regularization        & 0.7092 & 26.21 & 20.64 & 0.0715     \\ \hline
			& SS-TV    & 0.6830 & 23.81 & 18.24 & 0.0942    \\ \hline
			& Tikhonov & 0.3041 & 21.09 & 15.52 & 0.1289     \\ \hlineB{3}
			0.7      & Proposed & \textbf{0.7096} & \textbf{24.87} & \textbf{19.30} & \textbf{0.0834}     \\ \hline
			& $\ell^1$ regularization        & 0.6221 & 21.93 & 16.36 & 0.1170      \\ \hline
			& SS-TV    & 0.6816 & 22.60 & 17.03 & 0.1083      \\ \hline
			& Tikhonov & 0.1414 & 15.51 & 9.951 & 0.2445   \\ \hline
		\end{tabular}
	\end{table}

We can observe from \cref{table_all} that our proposed PCM-TV has a better performance than all the other three one-stage methods.

\begin{figure}[htbp]
		\centering
			\subfloat[Ground Truth]{\includegraphics[width=0.3\columnwidth]{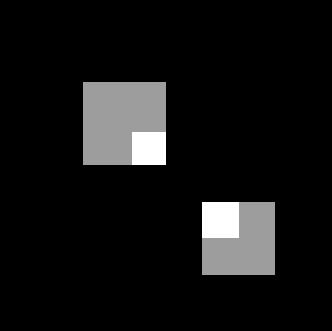}}
						\hspace{0.2in}
			\subfloat[SS-TV]{\includegraphics[width=0.3\columnwidth]{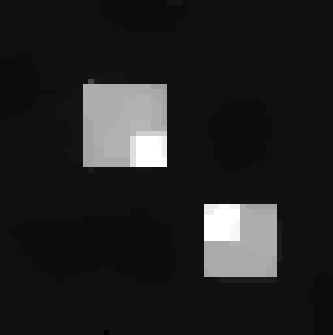}}
						\hspace{0.2in}
			\subfloat[PCM-TV]{\includegraphics[width=0.3\columnwidth]{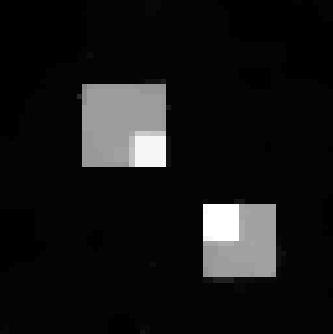}}

		\caption{ Reconstruction comparison with bias error in for 2D image. Here $m = 128 \times 128$, $n = 64\times 64$, $imsize = 256 \times 256$, $\sigma = 0.6$.}\label{2D_bias_all}
	\end{figure}

We next consider the 2D case. We will consider the 2D function $f$ with a randomly chosen square support of $[0.25,0.5]^2 \cup [0.61,0.83]^2$ under where the entire image region is defined as $[0,1]^2$. We display it in \Cref{2D_bias_all}. The 2D Haar wavelet used here is a direct product of 1D Haar wavelet leading to $\Hil_{n}^2 := \Hil_{n}\otimes \Hil_{n}$.

 In particular, we will use a piecewise constant test function whose support does not lie in $\Hil_n^2$. It consists of four squares and the white ones do not lie in $\Hil_n$, which will cause the bias error in reconstruction. 

We will compare the proposed PCM-TV model with the single stage TV (SS-TV) method. We point out that TV regularization usually yield better results than $\ell^1$ regularization and $\ell^2$ regularization in 2D imaging problem. We present the corresponding numerical results in \Cref{2D_table_bias_all}. 	
	\begin{table}[htbp]
		\centering
		\captionsetup{position=top}
		\caption{Numerical details for the 2D reconstruction in \cref{2D_bias_all} with bias error, $\sigma = 0.6$.}\label{2D_table_bias_all}
		\begin{tabular}{|l|l|l|l|}
			\hlineB{3}
			Model    & psnr  & snr   & rela\_err \\  \hlineB{3}
			SS-TV    & 33.8196 & 20.87 & 0.086     \\ \hline
			PCM-TV  &\textbf{35.6589} &\textbf{22.7092} & \textbf{0.069 }   \\ \hlineB{3}
		\end{tabular}
	\end{table}

We could observe from the above numerical results that the two stage PCM-TV has a better performance than the single stage TV regularization method. We would next explain the differences between them. We point out that the two stage PCM-TV model is equivalent to the following optimization problem,
\[
\min_{\vc_g} \frac{1}{2}\|\vc_g - A^\dagger\hat{f}\|^2 + \lambda\mathcal{R}(\vc_g),
\]
where $A = \mathcal{S}\F \Phi$ and $\cR$ is the TV operator. 
The corresponding first order optimality condition (from Fermat's rule \cite{Bauschke2011}) implies that
\begin{equation}\label{eq:Two}
\mathbf{0} \in \vc_g - A^\dagger\hat{f}+\lambda\partial\mathcal{R}(\vc_g),  
\end{equation}
where $\partial\mathcal{R}$ denotes the sub-differential of $\cR$. On the other hand, for the single stage SS-TV model 
\[
\min_{\vc_g} \frac{1}{2}\|A\vc_g - \hat{f}\|^2 + \lambda\mathcal{R}(\vc_g),
\]
the first order condition implies that
\begin{equation}\label{eq:One}
\mathbf{0} \in \vc_g - A^\dagger\hat{f}+\lambda(A^*A)^{-1}\partial\mathcal{R}(\vc_g).  
\end{equation}
Compared with \cref{eq:Two}, the descent direction in \cref{eq:One} is distorted by the factor $(A^*A)^{-1}$. It might not only cause extra computational cost, but also bring numerical instability and extra errors in computing the inverse of $A^*A$.

\subsection{Performance of TV-TFV regularity}
In this subsection we demonstrate the advantages of the proposed TV-TFV regularity. We consider the following general image denoising problem
	\[
	\min_u \frac{1}{2}\|u - \hat{u}\|_2^2 + \lambda \mathcal{R}(u),
	\]
	where $\hat{u}$ is the given noisy image. We will compare these three different regularities in the above model: TV, TFV, and TV-TFV.

\begin{figure}[htbp]
		\centerline{
			\subfloat[Noisy]{\includegraphics[width=0.28\columnwidth]{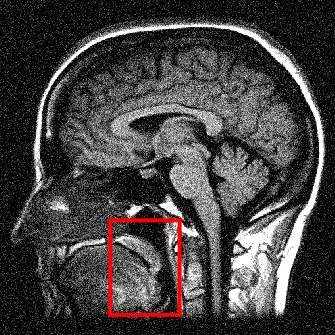}}
			\hspace{0.2in}
			\subfloat[Anisotropic TV]{\includegraphics[width=0.28\columnwidth]{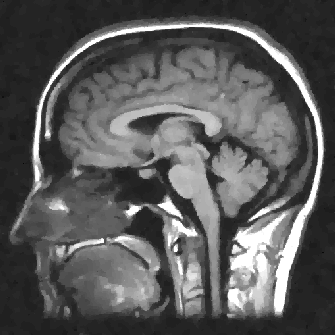}}
		}
		\centerline{
			\subfloat[TFV, $\alpha = 1.3$]{\includegraphics[width=0.28\columnwidth]{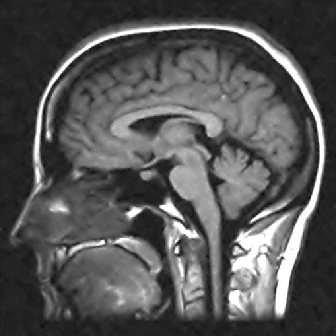}}
			\hspace{0.2in}
			\subfloat[TV-TFV, $\alpha = 1.3$]{\includegraphics[width=0.28\columnwidth]{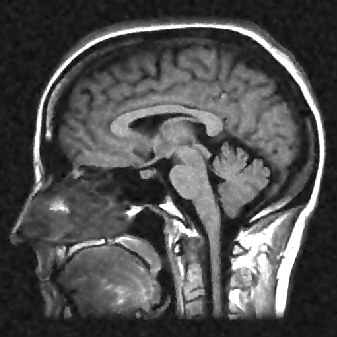}}
		}
		\caption{Denoising results for models with different regularities. Here $\alpha$ denotes the fractional order of derivative in TFV model.} \label{brain_denoising}
	\end{figure}

\begin{figure}[tbhp]
		\centerline{
			\subfloat[Ground Truth]{\includegraphics[width=0.28\columnwidth]{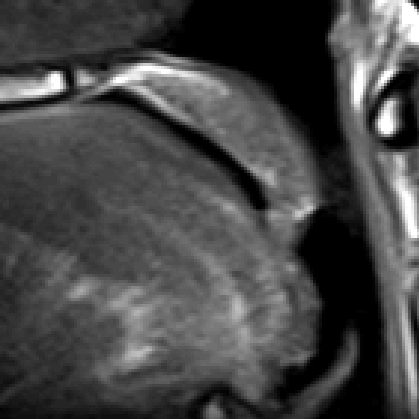}}
			\hspace{0.2in}
			\subfloat[Anisotropic TV]{\includegraphics[width=0.28\columnwidth]{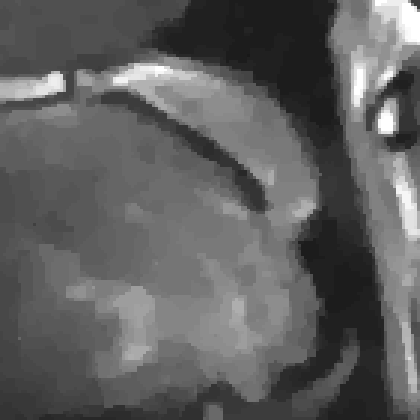}}
		}
		\centerline{
			\subfloat[TFV]{\includegraphics[width=0.28\columnwidth]{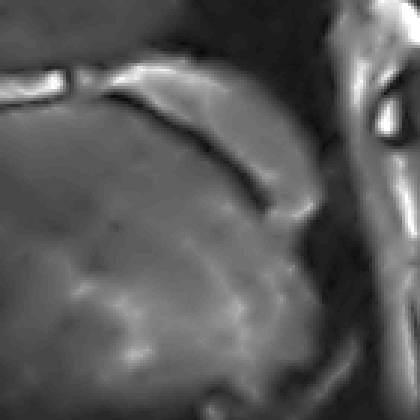}}
			\hspace{0.2in}
			\subfloat[TV-TFV]{\includegraphics[width=0.28\columnwidth]{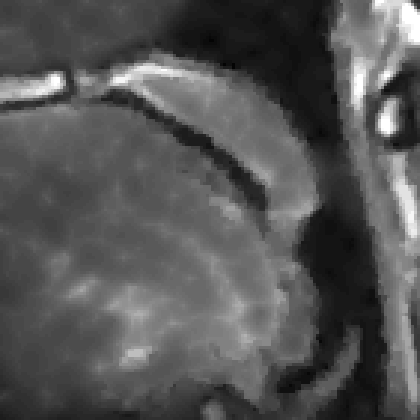}}
		}
		\caption{Detailed Comparison between different models.}\label{edges}
	\end{figure}

We display the noisy image and the reconstructed images from these three denoising methods in  \cref{brain_denoising} \footnote{	Image retrieved from \url{http://radiopaedia.org/} by Frank Gaillard.}. To better understand the performance, we zoom in the selected part of the image and display them in \cref{edges}. 
	
We observe that the anisotropic TV suffers from the staircase artifact due to the fact that the TV is local operator. On the other hand, the reconstruction with TFV regularity has blurry effect on the edges. This is not surprising because the TFV is a non-local method and it is less edge sensitive than TV. Instead, the TV-TFV regularity avoids such artifacts and has a better reconstruction of both the edges and the overall image. 

We also present the numerical results of different performance measurements in \cref{table_2d_denoising}. The TV-TFV regularity shows better results in such measurements as well. 
		\begin{table}[htbp]
			\captionsetup{position=top}
			\caption{Numerical results for denoising with different regularities. } \label{table_2d_denoising}
			\centering
			\begin{tabular}{|l|l|l|l|}
				\hlineB{3}
				Model    & psnr  & snr   & rela\_err \\  \hlineB{3}
				Noisy     &  20.9199 & 6.0853 &  0.3914  \\ \hline
				TV       & 31.0007 & 16.1661 & 0.1226   \\ \hline
				TFV     & 31.6910 & 16.8565 & 0.1133      \\ \hline
				TV-TFV  &\textbf{32.1948} &\textbf{17.3602} & \textbf{0.1069  }   \\ \hlineB{3}
			\end{tabular}
		\end{table}

\subsection{PCM-TV-TFV vs. PCM-TV}
Finally, we will combine the projection stage and the correction stage with TV-TFV regularity instead of TV regularity to further improve the performance. 

\begin{figure}[htbp]
		\centerline{
			\subfloat[Ground Truth]{\includegraphics[width=0.35\columnwidth]{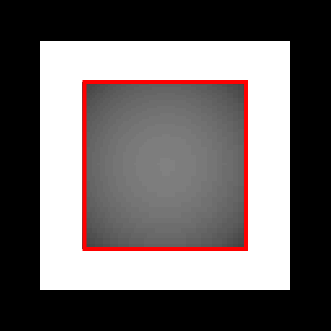}}
						\hspace{0.2in}
			\subfloat[Noisy]{\includegraphics[width=0.35\columnwidth]{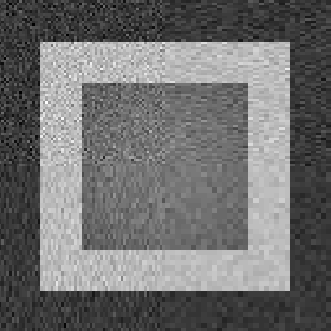}}
		}
		\centerline{
			\subfloat[PCM-TV]{\includegraphics[width=0.35\columnwidth]{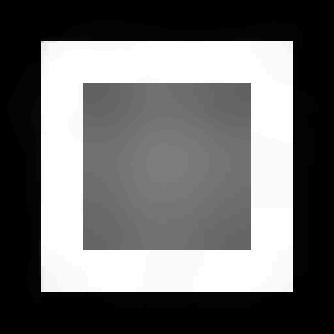}}
						\hspace{0.2in}
			\subfloat[PCM-TV-TFV]{\includegraphics[width=0.35\columnwidth]{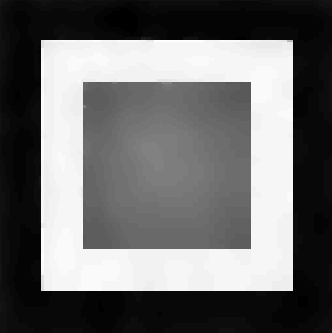}}
		} 
		\caption{Reconstruction comparison between PCM-TV and PCM-TV-TFV. Noisy image (b) is obtained from inverse Fourier transform. Here $m = 128 \times 128$, $n = 96\times 96$, $imsize = 256 \times 256$, $\sigma = 0.4$.}\label{2D_TV-TFV_vs_TV}
	\end{figure}

We display the ground truth image and the reconstructed images from the inverse Fourier method, the PCM-TV method and the PCM-TV-TFV method in \cref{2D_TV-TFV_vs_TV}.

\begin{figure}[htbp]
    	\centering
    	\subfloat[Groud Truth]{
    		\includegraphics[width=0.32\columnwidth]{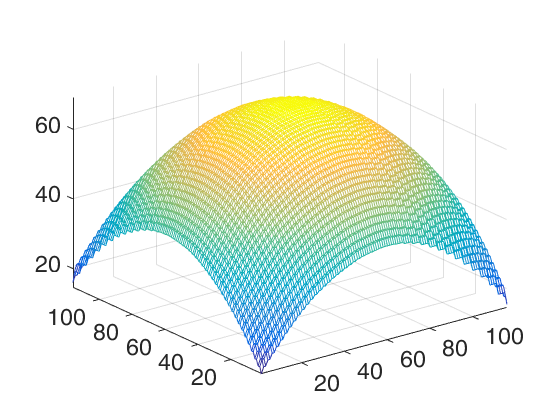}}
    	\subfloat[PCM-TV]{
    		\includegraphics[width=0.32\columnwidth]{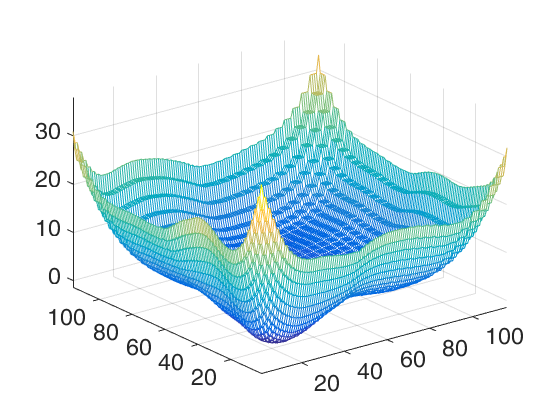}}
    	\subfloat[PCM-TV-TFV]{
    		\includegraphics[width=0.32\columnwidth]{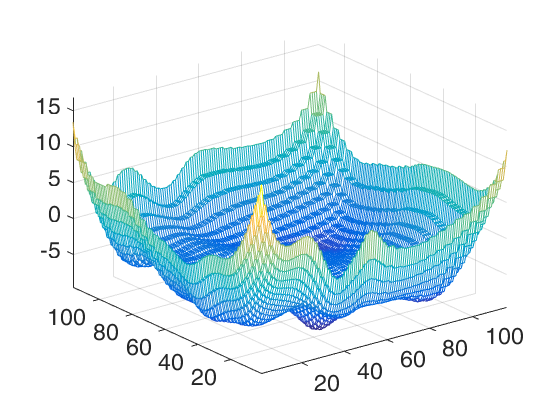}}
    	\caption{Zoomed in comparison of the red square in  \cref{2D_TV-TFV_vs_TV}. Figure (a) is the true surface. Figure (b) shows the difference between the truth and the one reconstructed by PCM-TV. Relative error $\approx 62\%$. Figure (c) shows the difference between the truth and the one reconstructed by PCM-TV-TFV. Relative error $\approx 32\%$.} \label{surface_TV-TFV_TV}
    \end{figure}

We point out that the noisy image in  \cref{2D_TV-TFV_vs_TV} is obtained directly by inverse Fourier transform and we can see that the noise level is quite high in this case. Both the PCM-TV and the PCM-TV-TFV are able to produce more reasonable visual results. To see a deep comparison, we zoom in the red square part of  \cref{2D_TV-TFV_vs_TV} and present the approximation errors in \cref{surface_TV-TFV_TV}.

We can see that the PCM-TV-TFV has a much less error than the PCM-TV model for the surface reconstruction.

\section{Conclusions}\label{conclusions}
In this paper we propose a general two stage Projection Correction Modeling framework for image reconstruction from Fourier measurements. The projection step alleviates the instability from the non-uniformness of the Fourier measurements and the correction step further reduces the noise and the bias effects in the previous step. A precise edge guided TV-TFV regularity shows its own advantages over the models with single TV or TFV regularity. The numerical experiments demonstrate that such combination enhances the reconstruction and reduces the drawbacks of the TV and TFV themselves. Furthermore, we also show that the proposed PCM-TV-TFV has a superior performance even when the measurements have considerably heavy noise. 

We remark that we only use wavelets to demonstrate the advantages of the proposed two stage PCM method in the numerical experiments. However, other basis such as shearlets, curvelets or adaptive wavelets could also be incorporated to further improve the results. Moreover, the proposed two stage framework for Fourier measurements could also be extended to other linear measurements such as Radon transform \etc.

\section*{Acknowledgment}
The authors would like to thank the discussion with Yiqiu Dong.

\bibliographystyle{siamplain}
\bibliography{PCM-SIAM}

\begin{thebibliography}{10}

\bibitem{adcock_generalized_2015}
{\sc B.~Adcock and A.~C. Hansen}, {\em Generalized {Sampling} and
  {Infinite}-{Dimensional} {Compressed} {Sensing}}, Found Comput Math,  (2015),
  pp.~1--61.

\bibitem{adcock2013breaking}
{\sc B.~Adcock, A.~C. Hansen, C.~Poon, and B.~Roman}, {\em Breaking the
  coherence barrier: A new theory for compressed sensing}, arXiv preprint
  arXiv:1302.0561,  (2013).

\bibitem{alonso_novel_2010}
{\sc M.~T. Alonso, P.~Lopez-Dekker, and J.~J. Mallorqui}, {\em A {Novel}
  {Strategy} for {Radar} {Imaging} {Based} on {Compressive} {Sensing}}, IEEE
  Transactions on Geoscience and Remote Sensing, 48 (2010), pp.~4285--4295.

\bibitem{archibald2016image}
{\sc R.~Archibald, A.~Gelb, and R.~B. Platte}, {\em Image reconstruction from
  undersampled fourier data using the polynomial annihilation transform},
  Journal of Scientific Computing, 67 (2016), pp.~432--452.

\bibitem{Bauschke2011}
{\sc H.~H. Bauschke and P.~L. Combettes}, {\em Convex analysis and monotone
  operator theory in {H}ilbert spaces}, CMS Books in Mathematics/Ouvrages de
  Math\'ematiques de la SMC, Springer, New York, 2011,
  \url{https://doi.org/10.1007/978-1-4419-9467-7},
  \url{http://dx.doi.org/10.1007/978-1-4419-9467-7}.
\newblock With a foreword by H{\'e}dy Attouch.

\bibitem{beck2009fast}
{\sc A.~Beck and M.~Teboulle}, {\em A fast iterative shrinkage-thresholding
  algorithm for linear inverse problems}, SIAM journal on imaging sciences, 2
  (2009), pp.~183--202.

\bibitem{becker_nesta:_2011}
{\sc S.~Becker, J.~Bobin, and E.~J. Candès}, {\em {NESTA}: a fast and accurate
  first-order method for sparse recovery}, SIAM Journal on Imaging Sciences, 4
  (2011), pp.~1--39.

\bibitem{DBLP:journals/mpc/BeckerCG11}
{\sc S.~Becker, E.~J. Cand{\`{e}}s, and M.~C. Grant}, {\em Templates for convex
  cone problems with applications to sparse signal recovery}, Math. Program.
  Comput., 3 (2011), pp.~165--218.

\bibitem{boyd2011distributed}
{\sc S.~Boyd, N.~Parikh, E.~Chu, B.~Peleato, and J.~Eckstein}, {\em Distributed
  optimization and statistical learning via the alternating direction method of
  multipliers}, Foundations and Trends{\textregistered} in Machine Learning, 3
  (2011), pp.~1--122.

\bibitem{brankov2002multi}
{\sc J.~G. Brankov, Y.~Yang, R.~M. Leahy, and M.~N. Wernick}, {\em
  Multi-modality tomographic image reconstruction using mesh modeling}, in
  Biomedical Imaging, 2002. Proceedings. 2002 IEEE International Symposium on,
  IEEE, 2002, pp.~405--408.

\bibitem{bredies2010total}
{\sc K.~Bredies, K.~Kunisch, and T.~Pock}, {\em Total generalized variation},
  SIAM Journal on Imaging Sciences, 3 (2010), pp.~492--526.

\bibitem{cai2016image}
{\sc J.-F. Cai, B.~Dong, and Z.~Shen}, {\em Image restoration: a wavelet frame
  based model for piecewise smooth functions and beyond}, Applied and
  Computational Harmonic Analysis, 41 (2016), pp.~94--138.

\bibitem{CPA:CPA21455}
{\sc E.~J. Candès and C.~Fernandez-Granda}, {\em Towards a mathematical theory
  of super-resolution}, Communications on Pure and Applied Mathematics, 67
  (2014), pp.~906--956.

\bibitem{canny1986computational}
{\sc J.~Canny}, {\em A computational approach to edge detection}, IEEE
  Transactions on pattern analysis and machine intelligence,  (1986),
  pp.~679--698.

\bibitem{chan2000high}
{\sc T.~Chan, A.~Marquina, and P.~Mulet}, {\em High-order total variation-based
  image restoration}, SIAM Journal on Scientific Computing, 22 (2000),
  pp.~503--516.

\bibitem{chen2013fractional}
{\sc D.~Chen, Y.~Chen, and D.~Xue}, {\em Fractional-order total variation image
  restoration based on primal-dual algorithm}, in Abstract and Applied
  Analysis, vol.~2013, Hindawi Publishing Corporation, 2013.

\bibitem{chen2013fractional2}
{\sc D.~Chen, S.~Sun, C.~Zhang, Y.~Chen, and D.~Xue}, {\em Fractional-order
  tv-l2 model for image denoising}, Central European Journal of Physics, 11
  (2013), pp.~1414--1422.

\bibitem{chen_time-frequency_2002}
{\sc V.~C. Chen and H.~Ling}, {\em Time-frequency transforms for radar imaging
  and signal analysis}, Artech House, 2002.

\bibitem{chernyakova_fourier-domain_2014}
{\sc T.~Chernyakova and Y.~C. Eldar}, {\em Fourier-domain beamforming: the path
  to compressed ultrasound imaging}, IEEE transactions on ultrasonics,
  ferroelectrics, and frequency control, 61 (2014), pp.~1252--1267.

\bibitem{choi2017edge}
{\sc J.~K. Choi, B.~Dong, and X.~Zhang}, {\em An edge driven wavelet frame
  model for image restoration}, arXiv preprint arXiv:1701.07158,  (2017).

\bibitem{chumchob2011fourth}
{\sc N.~Chumchob, K.~Chen, and C.~Brito-Loeza}, {\em A fourth-order variational
  image registration model and its fast multigrid algorithm}, Multiscale
  Modeling \& Simulation, 9 (2011), pp.~89--128.

\bibitem{cui2014total}
{\sc X.~Cui, H.~Yu, G.~Wang, and L.~Mili}, {\em Total variation
  minimization-based multimodality medical image reconstruction}, in SPIE
  Optical Engineering+ Applications, International Society for Optics and
  Photonics, 2014, pp.~92121D--92121D.

\bibitem{davies2012computer}
{\sc E.~R. Davies}, {\em Computer and machine vision: theory, algorithms,
  practicalities}, Academic Press, 2012.

\bibitem{deng2016single}
{\sc L.-J. Deng, W.~Guo, and T.-Z. Huang}, {\em Single-image super-resolution
  via an iterative reproducing kernel hilbert space method}, IEEE Transactions
  on Circuits and Systems for Video Technology, 26 (2016), pp.~2001--2014.

\bibitem{dennis1996numerical}
{\sc J.~E. Dennis~Jr and R.~B. Schnabel}, {\em Numerical methods for
  unconstrained optimization and nonlinear equations}, SIAM, 1996.

\bibitem{dong_mra-based_2012}
{\sc B.~Dong and Z.~Shen}, {\em {MRA}-based wavelet frames and applications:
  {Image} segmentation and surface reconstruction}, in {SPIE} {Defense},
  {Security}, and {Sensing}, International Society for Optics and Photonics,
  2012, pp.~840102--840102.

\bibitem{easley2008sparse}
{\sc G.~Easley, D.~Labate, and W.-Q. Lim}, {\em Sparse directional image
  representations using the discrete shearlet transform}, Applied and
  Computational Harmonic Analysis, 25 (2008), pp.~25--46.

\bibitem{gelb_frame_2014}
{\sc A.~Gelb and G.~Song}, {\em A {Frame} {Theoretic} {Approach} to the
  {Nonuniform} {Fast} {Fourier} {Transform}}, SIAM J. Numer. Anal., 52 (2014),
  pp.~1222--1242.

\bibitem{gockenbach2006understanding}
{\sc M.~S. Gockenbach}, {\em Understanding and implementing the finite element
  method}, Siam, 2006.

\bibitem{goldstein2014fast}
{\sc T.~Goldstein, B.~O'Donoghue, S.~Setzer, and R.~Baraniuk}, {\em Fast
  alternating direction optimization methods}, SIAM Journal on Imaging
  Sciences, 7 (2014), pp.~1588--1623.

\bibitem{goldstein2009split}
{\sc T.~Goldstein and S.~Osher}, {\em The split bregman method for
  l1-regularized problems}, SIAM journal on imaging sciences, 2 (2009),
  pp.~323--343.

\bibitem{gorenflo1997fractional}
{\sc R.~Gorenflo and F.~Mainardi}, {\em Fractional calculus}, Springer, 1997.

\bibitem{griswold2002generalized}
{\sc M.~A. Griswold, P.~M. Jakob, R.~M. Heidemann, M.~Nittka, V.~Jellus,
  J.~Wang, B.~Kiefer, and A.~Haase}, {\em Generalized autocalibrating partially
  parallel acquisitions (grappa)}, Magnetic resonance in medicine, 47 (2002),
  pp.~1202--1210.

\bibitem{guo2014new}
{\sc W.~Guo, J.~Qin, and W.~Yin}, {\em A new detail-preserving regularization
  scheme}, SIAM Journal on Imaging Sciences, 7 (2014), pp.~1309--1334.

\bibitem{guo2012edge}
{\sc W.~Guo and W.~Yin}, {\em Edge guided reconstruction for compressive
  imaging}, SIAM Journal on Imaging Sciences, 5 (2012), pp.~809--834.

\bibitem{he2007mr}
{\sc L.~He, T.-C. Chang, S.~Osher, T.~Fang, and P.~Speier}, {\em Mr image
  reconstruction from undersampled data by using the iterative refinement
  procedure}, Pamm, 7 (2007), pp.~1011207--1011208.

\bibitem{james2014medical}
{\sc A.~P. James and B.~V. Dasarathy}, {\em Medical image fusion: A survey of
  the state of the art}, Information Fusion, 19 (2014), pp.~4--19.

\bibitem{knoll2011second}
{\sc F.~Knoll, K.~Bredies, T.~Pock, and R.~Stollberger}, {\em Second order
  total generalized variation (tgv) for mri}, Magnetic resonance in medicine,
  65 (2011), pp.~480--491.

\bibitem{li2015multiphase}
{\sc F.~Li, S.~Osher, J.~Qin, and M.~Yan}, {\em A multiphase image segmentation
  based on fuzzy membership functions and l1-norm fidelity}, Journal of
  Scientific Computing,  (2015), pp.~1--25.

\bibitem{lu2014edge}
{\sc Y.~Lu, J.~Zhao, and G.~Wang}, {\em Edge-guided dual-modality image
  reconstruction}, IEEE Access, 2 (2014), pp.~1359--1363.

\bibitem{lustig_sparse_2007}
{\sc M.~Lustig, D.~Donoho, and J.~M. Pauly}, {\em Sparse {MRI}: {The}
  application of compressed sensing for rapid {MR} imaging}, Magnetic resonance
  in medicine, 58 (2007), pp.~1182--1195.

\bibitem{lustig_compressed_2008}
{\sc M.~Lustig, D.~L. Donoho, J.~M. Santos, and J.~M. Pauly}, {\em Compressed
  sensing {MRI}}, IEEE Signal Processing Magazine, 25 (2008), pp.~72--82.

\bibitem{ma2010curvelet}
{\sc J.~Ma and G.~Plonka}, {\em The curvelet transform}, IEEE Signal Processing
  Magazine, 27 (2010), pp.~118--133.

\bibitem{melbourne2012using}
{\sc A.~Melbourne, N.~Cahill, C.~Tanner, M.~Modat, D.~Hawkes, and S.~Ourselin},
  {\em Using fractional gradient information in non-rigid image registration:
  application to breast mri}, in SPIE Medical Imaging, International Society
  for Optics and Photonics, 2012, pp.~83141Z--83141Z.

\bibitem{miller1993introduction}
{\sc K.~S. Miller and B.~Ross}, {\em An introduction to the fractional calculus
  and fractional differential equations}, 1993.

\bibitem{nesterov1983method}
{\sc Y.~Nesterov}, {\em A method of solving a convex programming problem with
  convergence rate o (1/k2)}, in Soviet Mathematics Doklady, vol.~27, 1983,
  pp.~372--376.

\bibitem{osher2005iterative}
{\sc S.~Osher, M.~Burger, D.~Goldfarb, J.~Xu, and W.~Yin}, {\em An iterative
  regularization method for total variation-based image restoration},
  Multiscale Modeling \& Simulation, 4 (2005), pp.~460--489.

\bibitem{osherfast}
{\sc S.~Osher, Y.~Mao, B.~Dong, and W.~Yin}, {\em Fast linearized bregman
  iteration for compressive sensing and sparse denoising}, Communications in
  Mathematical Sciences, 8 (2010), pp.~93--111.

\bibitem{parikh2014proximal}
{\sc N.~Parikh, S.~P. Boyd, et~al.}, {\em Proximal algorithms.}, Foundations
  and Trends in optimization, 1 (2014), pp.~127--239.

\bibitem{podlubny2000matrix}
{\sc I.~Podlubny}, {\em Matrix approach to discrete fractional calculus},
  Fractional Calculus and Applied Analysis, 3 (2000), pp.~359--386.

\bibitem{pruessmann1999sense}
{\sc K.~P. Pruessmann, M.~Weiger, M.~B. Scheidegger, P.~Boesiger, et~al.}, {\em
  Sense: sensitivity encoding for fast mri}, Magnetic resonance in medicine, 42
  (1999), pp.~952--962.

\bibitem{rank1999estimation}
{\sc K.~Rank, M.~Lendl, and R.~Unbehauen}, {\em Estimation of image noise
  variance}, IEE Proceedings-Vision, Image and Signal Processing, 146 (1999),
  pp.~80--84.

\bibitem{rudin1992nonlinear}
{\sc L.~I. Rudin, S.~Osher, and E.~Fatemi}, {\em Nonlinear total variation
  based noise removal algorithms}, Physica D: Nonlinear Phenomena, 60 (1992),
  pp.~259--268.

\bibitem{serra1982image}
{\sc J.~Serra}, {\em Image analysis and mathematical morphology, v. 1},
  Academic press, 1982.

\bibitem{song2013approximating}
{\sc G.~Song and A.~Gelb}, {\em Approximating the inverse frame operator from
  localized frames}, Applied and Computational Harmonic Analysis, 35 (2013),
  pp.~94--110.

\bibitem{starck2002curvelet}
{\sc J.-L. Starck, E.~J. Cand{\`e}s, and D.~L. Donoho}, {\em The curvelet
  transform for image denoising}, IEEE Transactions on image processing, 11
  (2002), pp.~670--684.

\bibitem{szeliski2010computer}
{\sc R.~Szeliski}, {\em Computer vision: algorithms and applications}, Springer
  Science \& Business Media, 2010.

\bibitem{verdu2009fractional}
{\sc R.~Verd{\'u}-Monedero, J.~Larrey-Ruiz, J.~Morales-S{\'a}nchez, and J.~L.
  Sancho-G{\'o}mez}, {\em Fractional regularization term for variational image
  registration}, Mathematical Problems in Engineering, 2009 (2009).

\bibitem{wagner_compressed_2012}
{\sc N.~Wagner, Y.~C. Eldar, and Z.~Friedman}, {\em Compressed beamforming in
  ultrasound imaging}, IEEE Transactions on Signal Processing, 60 (2012),
  pp.~4643--4657.

\bibitem{wahlberg_admm_2012}
{\sc B.~Wahlberg, S.~Boyd, M.~Annergren, and Y.~Wang}, {\em An {ADMM} algorithm
  for a class of total variation regularized estimation problems}, IFAC
  Proceedings Volumes, 45 (2012), pp.~83--88.

\bibitem{wang2004image}
{\sc Z.~Wang, A.~C. Bovik, H.~R. Sheikh, and E.~P. Simoncelli}, {\em Image
  quality assessment: from error visibility to structural similarity}, IEEE
  transactions on image processing, 13 (2004), pp.~600--612.

\bibitem{yin2008bregman}
{\sc W.~Yin, S.~Osher, D.~Goldfarb, and J.~Darbon}, {\em Bregman iterative
  algorithms for $\backslash$ell\_1-minimization with applications to
  compressed sensing}, SIAM Journal on Imaging sciences, 1 (2008),
  pp.~143--168.

\bibitem{zhang2015total}
{\sc J.~Zhang and K.~Chen}, {\em A total fractional-order variation model for
  image restoration with nonhomogeneous boundary conditions and its numerical
  solution}, SIAM Journal on Imaging Sciences, 8 (2015), pp.~2487--2518.

\bibitem{zhang__2013}
{\sc Y.~Zhang, B.~Dong, and Z.~Lu}, {\em ℓ₀ {Minimization} for wavelet
  frame based image restoration}, Mathematics of Computation, 82 (2013),
  pp.~995--1015.

\bibitem{zhang2012class}
{\sc Y.~Zhang, Y.~Pu, J.~Hu, and J.~Zhou}, {\em A class of fractional-order
  variational image inpainting models}, Appl. Math. Inf. Sci, 6 (2012),
  pp.~299--306.

\bibitem{zulfiquar_ali_bhotto_improved_2015}
{\sc M.~Zulfiquar Ali~Bhotto, M.~O. Ahmad, and M.~N.~S. Swamy}, {\em An
  {Improved} {Fast} {Iterative} {Shrinkage} {Thresholding} {Algorithm} for
  {Image} {Deblurring}}, SIAM Journal on Imaging Sciences, 8 (2015),
  pp.~1640--1657.

\end{thebibliography}

\end{document}